\documentclass[conference]{IEEEtran}
\usepackage{times}

\usepackage[numbers]{natbib}
\usepackage{multicol}
\usepackage[bookmarks=true]{hyperref}
\usepackage{graphicx}
\usepackage{amsmath}
\usepackage{cuted} 
\usepackage{xcolor} 
\usepackage{capt-of}  

\usepackage{color, soul} 

\author{M. Nomaan Qureshi, Ben Eisner, David Held\\

Carnegie Mellon University
}

\begin{document}

\title{On Time-Indexing as Inductive Bias in Deep RL for Sequential Manipulation Tasks}

\maketitle

\begin{abstract}
While solving complex manipulation tasks, manipulation policies often need to learn a set of diverse skills to accomplish these tasks. The set of skills is often quite multimodal - each one may have a quite distinct distribution of actions and states. Standard deep policy-learning algorithms often model policies as deep neural networks with a single output head (deterministic or stochastic). This structure requires the network to learn to switch between modes internally, which can lead to lower sample efficiency and poor performance. In this paper we explore a simple structure which is conducive to skill learning required for so many of the manipulation tasks. Specifically, we propose a policy architecture that sequentially executes different action heads for fixed durations, enabling the learning of primitive skills such as reaching and grasping. Our empirical evaluation on the Metaworld tasks reveals that this simple structure outperforms standard policy learning methods, highlighting its potential for improved skill acquisition. 
\end{abstract}

\IEEEpeerreviewmaketitle

\section{Introduction}


\begin{figure}[h]
\begin{center}
\includegraphics[width=0.5\textwidth]{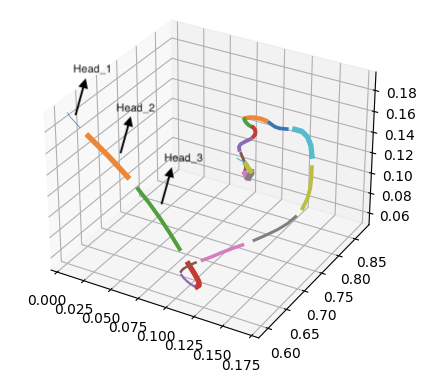}
\end{center}
  \caption{A visual representation of our policy. The action heads are sequentially executed for a fixed amount of time. This gives policy a structure conducive for skill learning. The policy can utilise these heads to learn primitive skills as reach, grasp etc. }
\label{teaser}
\end{figure}

 Consider a robotic manipulation task that involves stacking a book on a shelf. To achieve this objective, a robot must undertake a series of steps, such as reaching for the book, grasping it, identifying a suitable location for placement, and finally placing the book on the shelf. Although the location of the book and the relative positioning of the book and shelf may vary, the fundamental skills required to perform the task remain constant. These skills can be sequentially executed to accomplish the desired outcome. In the context of quasi-static manipulation tasks, a specific sequence of skills, when performed consecutively, can lead to successful task completion. How can we induce a similar structure to the modern neural-network based policies?


In the context of the previously mentioned example, it is evident that the requisite skills for executing the various sub-tasks are substantially different. For instance, the action distribution for the motor program necessary for grasping an object likely differs substantially from that required for reaching an object. In standard policy learning, a single neural-network based policy is tasked with learning both of these skills (and learning to switch between them),  without any access to structures that explicitly encode the multi-modal nature of task space.
Ideally, policies would be able to emergently learn to decompose tasks at different levels of abstraction, and factor the task learning into unique skills.


One common approach is to try and jointly learn a set of subskills, as well as a selection function which selects a specific subskill to execute at the current time step \cite{SUTTON1999181}. This poses a fundamental bootstrapping issue: as the skills change and improve, the selection function must change and improve as well, which can lead to unstable training. An important observation of many optimal policies for manipulation tasks is that skills tend to be executed in sequence, without backtracking. Therefore, \textit{time itself} can serve as a useful indicator for skill selection. For instance, while executing a stacking task, it is reasonable to assume that the robot will undertake the 'reach' skill at the start of the task, and subsequently perform the 'stack' skill towards the end of the task. Our intuition here is that selecting the 'skill' according to which time-step we are currently at can be used as a good strategy for selecting the skill to execute.  

In this paper we investigate how we can provide a neural network policy with a structure which is more conducive with skill learning. We particularly present a policy which switches the skills according to time and show that even this simple structure leads to better learning of manipulation tasks. Through these preliminary results we want to emphasise further research in this area. 


\begin{strip}
\begin{center}
\includegraphics[width=1.0\linewidth]{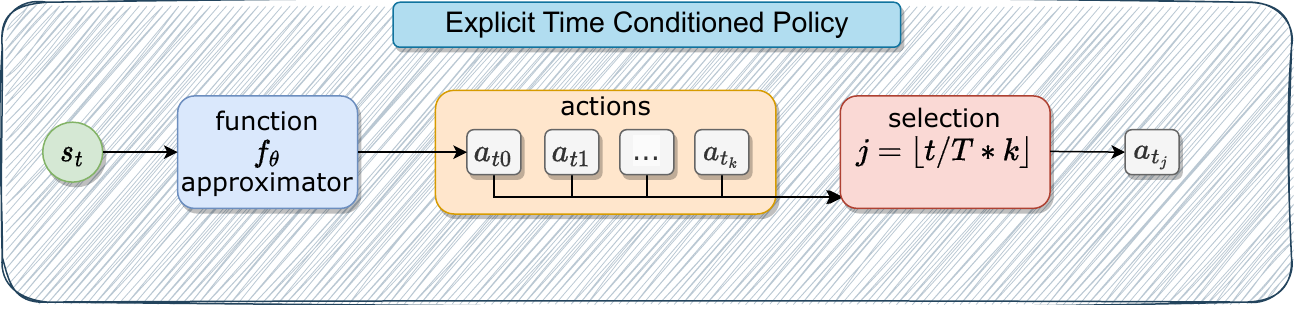}
\end{center}
  \captionof{figure}{A visual representation of the proposed algorithm depicting the time-indexed policy structure. The policy consists of multiple action heads that are sequentially activated for fixed durations. Each head corresponds to a specific skill or action, enabling the policy to learn specialized skills and integrate them to perform complex tasks.}
\label{pipeline}
\end{strip}

\section{Related Work}
Hierarchical reinforcement learning (HRL) \cite{SUTTON1999181, BaconHP16} represents a compelling area of investigation wherein skills are abstracted and organized into multiple hierarchical levels. Nonetheless, the application of HRL to complex environments presents major challenges, primarily stemming from the intricacy associated with concurrently learning policies at distinct hierarchical levels. In contrast, our method prescribes a simple time-indexed sequence, dodging the issue of learning hierarchies.
Another set of methods have recently explored learning skills from unstructured demonstration data \cite{shankar2020discovering,pmlr-v119-shankar20b}. However we focus on providing policy structure while learning online. Our inituition is similar to \cite{qureshi2022deep}, however, we investigate the effect of simply using neural heads instead of classical motion primitives.

\section{Preliminaries}

We consider the standard reinforcement learning setting, where an agent observes state $s_t$, chooses action $a_t$ sampled from $\pi_\theta(s_t)$, and receives reward $r_t$. The agent's objective is to maximize cumulative rewards in an episode. 
\begin{align}
\label{eq:prelim_1}
\textstyle \mathbf{J}^\pi(s_0)=  E_{a_t\sim \pi_{\theta(s_t)},{s_0\sim p_0}}[\sum_{t=1}^{t=T} {\gamma}^{t} r(s_t, a_t)]
 \end{align}



\section{Method}

 Contemporary deep reinforcement learning algorithms frequently represent policies with neural networks which are conditioned on the state of the environment. In this work, we seek to add a small transformation to the observation space, namely the time-step $t$. There are two ways to condition the policy on the input. One simple way is to just append $t$ to the current state $s_t$ to get a new state ${n}_t$. The action $a_t$ is now generated using the equation
\begin{align}
\label{eq:prelim_2}
n_t = \{ s_t, t \} \quad 
a_t = \pi_\theta(n_t)
 \end{align}
However this still places the responsibility on the policy to deduce the relation between time-step and the action. Moreover it doesn't provide the policy with any specific structure. 

In our method we propose to explicitly condition the output \textbf{selection} of the policy on the $t$. Our function approximator $f_\theta$ takes in the state $s_t$ and outputs $k$ different possible actions $(a_{t,0}, a_{t,1}, .., a_{t,k})$. 
\begin{align}
\label{eq:prelim_3}
(a_{t0}, a_{t1}, .., a_{tk}) = f_\theta(s_t) 
 \end{align}
At any given time-step $t$, we choose the action $a_{tj}$:
\begin{align}
\label{eq:prelim_4}
j &= \lfloor t/T \cdot k \rfloor 
\end{align}
where $T$ represents the maximum episode length. 
\begin{figure*}[h!]
\begin{center}
\includegraphics[width=0.48\textwidth]{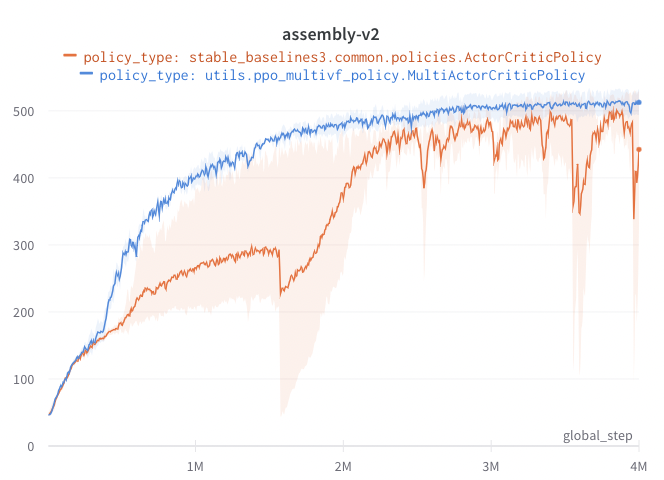}
\includegraphics[width=0.49\textwidth]{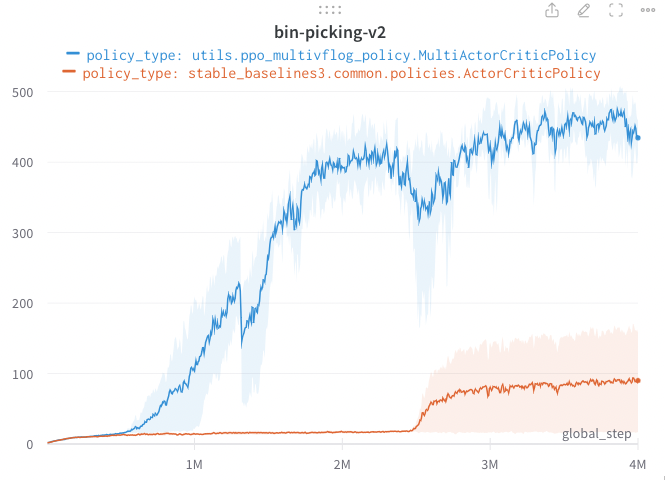}
\includegraphics[width=0.48\textwidth]{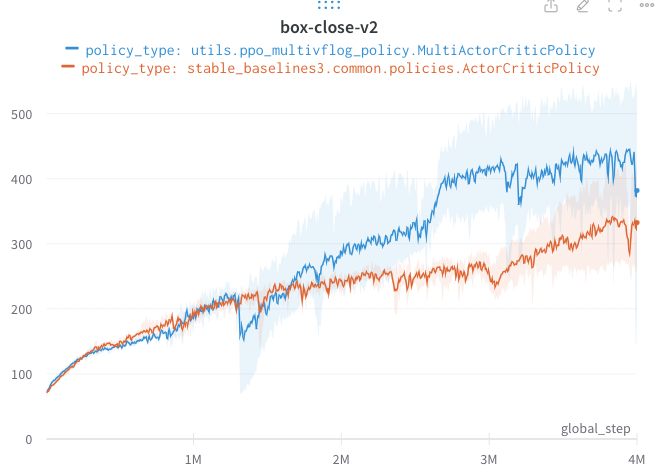}
\includegraphics[width=0.48\textwidth]{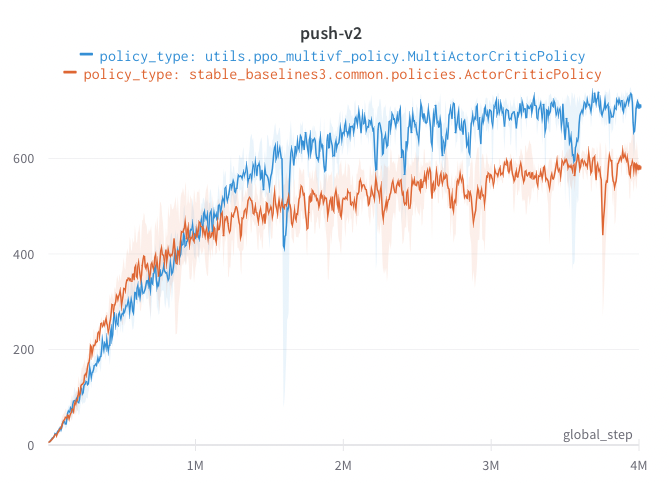}
\end{center}
  \caption{Primary Results : The figure illustrates the comparative performance of our algorithm against the standard implementation of Proximal Policy Optimization (PPO) on four different tasks. The plot showcases the average reward achieved as a function of the number of training steps. Our algorithm consistently outperforms the standard PPO across all tasks, demonstrating its effectiveness in discovering improved solutions and yielding higher rewards over time.}
\label{main_fig}
\vspace{-1.5em}
\end{figure*}
Intuitively, this means that the policy will output $k$ actions, and the current time-step of the episode us used to select which action to execute. For example, if our maximum episode length $T=100$ and number of heads $k=20$, so for the time-steps $t=\{1..5\}$, action from the first head, that is $a_{t0}$ will be used to collect the transitions $(s_t, a_{t0}, r_t, s_{t+1})$. For time-steps $t=\{6..10\}$, the actions from the second head $a_{t1}$ will be used to collect the transitions $(s_t, a_{t1}, r_t, s_{t+1})$ and so on. Each action head gets a fixed amount of time in which its action is applied on the environment. The heads are sequentially selected and remain active for a fixed amount of time. Fig. \ref{pipeline} visually describes our  algorithm.

The provides the policy with a structure which explicitly learns a sequence of skills to execute. Because the sequencing is explicit, each head of the network can learna  distinct skill mode without having to learn to switch between them.

\section{Experiments}
\subsection{\textbf{Environments}}
We evaluate the effectiveness of this simple policy change on four manipulation tasks from the MetaWorld environments\cite{yu2021metaworld}, including several sequential manipulation tasks. 
The tasks are : 
\begin{itemize}
    \item \texttt{assembly-v2}: In this task, the robot must assemble a simple structure using a set of parts. 
    \item \texttt{bin-picking-v2}: In this task, the robot must pick objects from a bin and place them in anther bin.
    \item \texttt{box-close-v2}: In this task, the robot must close a box by picking and placing its lid on top of the robot. 
    \item \texttt{push-v2}: In this task, the robot must push an object to a target location
\end{itemize}


\subsection{\textbf{Main results}}
Because our method is a simple modification of a neural network architecture, it can be integrated into most standard on-policy and off-policy RL algorithms such as Proximal Policy Optimization (PPO) and Soft Actor-Critic (SAC). Within this sub-section, a comparative evaluation is conducted, contrasting the performance of our algorithm with the standard implementation of PPO across the aforementioned tasks. Figure \ref{main_fig} presents a visual depiction of the average reward attained as a function of the number of training steps. Remarkably, our algorithm consistently outperforms the standard PPO approach across all four tasks.

Specifically, in tasks such as \texttt{push-v2} and \texttt{box-close-v2}, our algorithm significantly outperforms standard PPO, demonstrating the efficacy of explicit skill switching. Furthermore, in the case of \texttt{bin-picking-v2}, where the standard PPO fails to successfully solve the task, our multi-headed variant demonstrates efficient policy training. Finally, for the \texttt{assembly-v2} task, the multi-headed algorithm exhibits enhanced stability during training across multiple randomized seeds when contrasted with its standard counterpart. These empirical findings underscore the advantage of employing our algorithm in achieving superior performance in policy learning scenarios. 

\begin{figure}[!ht]
\begin{center}
\includegraphics[width=0.42\textwidth]{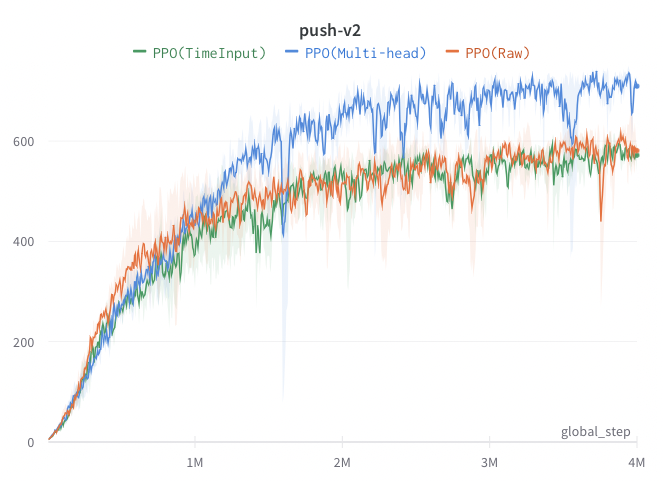}\\
\includegraphics[width=0.42\textwidth]{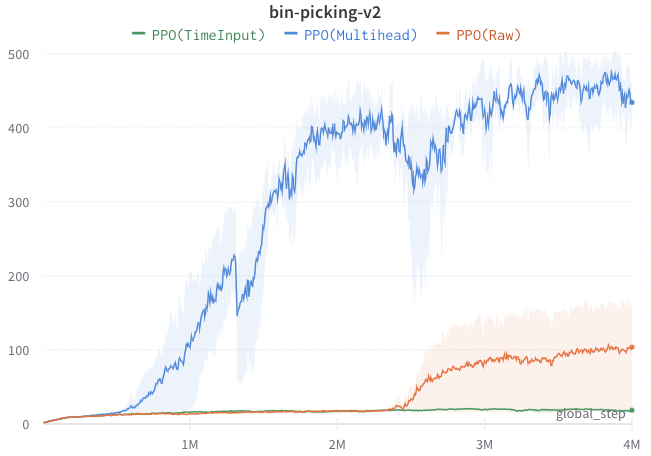}
\end{center}
\vspace{-1em}
  \caption{A comparison of using time to index policy heads (MultiHead) and including time in the observation.}
\label{abl1}
\end{figure}

Finally, in Fig. \ref{abl1} we compare our method with a policy which is provided directly as an input. We find that this formulation yields no benefit, as it adds no additional structure to the policy and suffers from the same multimodal modeling issue as a standard network.











\section{Conclusion} 
\label{sec:conclusion}
We present a simple inductive bias for use in reinforcement learning for robotic manipulation tasks by explicitly conditioning the policy on the current time step. We conduct empirical evaluations on various tasks, we have demonstrated improved performance of our algorithm over baselines in several sequential manipulation tasks. We anticipate that this research will inspire further exploration and development of structured policy designs for enhanced skill acquisition in robotic manipulation.

\begin{strip}
\begin{center}
\includegraphics[width=1.0\textwidth]{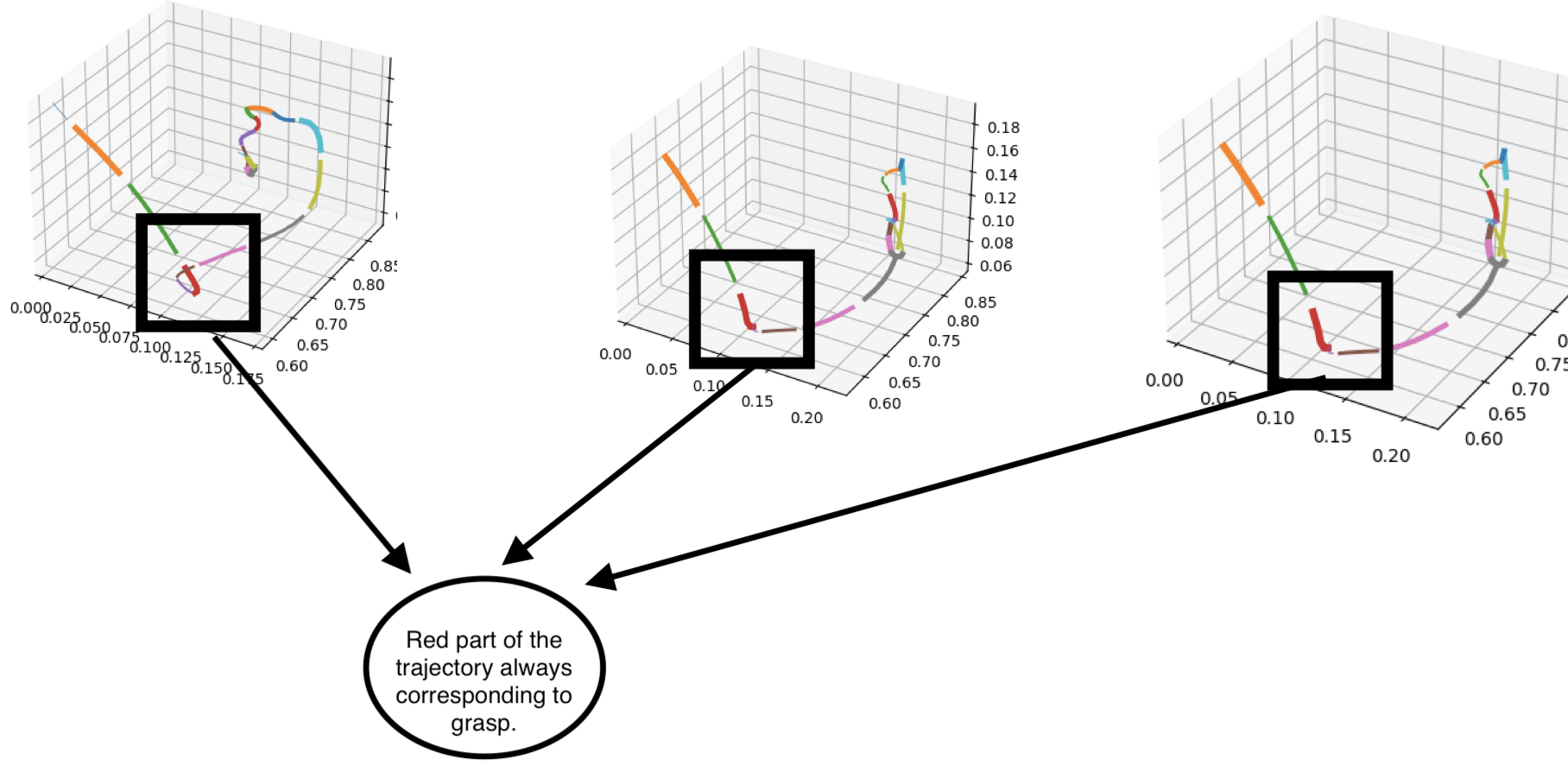}
\end{center}
\captionof{figure}{This figure shows the trajectory taken by the policy while attempting the assembly-v2 task. Each color represents the heads getting executed at different time-steps. We can see that the policy uses the different action heads to compose skills. For example the first two action heads (orange and green) are used to compose the reaching skill. The next two heads (red and purple)  are then used to compose grasping skill and so on.}
\label{pipeline2}
\end{strip}

\bibliographystyle{plainnat}
\bibliography{references}

\end{document}